\pdfoutput=1
\documentclass[11pt]{article}

\usepackage{acl} %for final version

\usepackage{times}
\usepackage{latexsym}

\usepackage[T1]{fontenc}
\usepackage[utf8]{inputenc}

\usepackage{microtype}

\usepackage{inconsolata}

\usepackage{graphicx}
\usepackage{booktabs}
\usepackage{multirow}
\usepackage{makecell}
\usepackage{amsmath}
\usepackage{xspace}
\usepackage{subcaption}
\usepackage{amsfonts}
\newcommand{\method}{\texttt{KNN-SSD}\xspace}
\newcommand{\ssdf}{\texttt{SSD\_F}\xspace}
\newcommand{\ssdm}{\texttt{SSD\_M}\xspace}

\title{\method: Enabling Dynamic Self-Speculative Decoding via\\Nearest Neighbor Layer Set Optimization}

\author{{Mingbo Song}\textsuperscript{\rm 1,3}, {\textbf{Heming Xia}}\textsuperscript{\rm 2}, {\textbf{Jun Zhang}}\textsuperscript{\rm 4}, {\textbf{Chak Tou Leong}}\textsuperscript{\rm 2},\\{\textbf{Qiancheng Xu}}\textsuperscript{\rm 2},{\textbf{Wenjie Li}}\textsuperscript{\rm 2}, {\textbf{Sujian Li}}\textsuperscript{\rm 1}\thanks{Corresponding author.}\\
  \textsuperscript{\rm 1}National Key Laboratory for Multimedia Information Processing, Peking University \\
  \textsuperscript{\rm 2}Department of Computing, The Hong Kong Polytechnic University \\
  \textsuperscript{\rm 3}Yuanpei College, Peking University \\
  \textsuperscript{\rm 4}College of Computer Science and Technology, Zhejiang University \\
  {\tt songmingbo@stu.pku.edu.cn; he-ming.xia@connect.polyu.hk;}\\
  {\tt lisujian@pku.edu.cn}}

\begin{document}
\maketitle
\begin{abstract}

Speculative Decoding (SD) has emerged as a widely used paradigm to accelerate the inference of large language models (LLMs) without compromising generation quality.
It works by efficiently drafting multiple tokens using a compact model and then verifying them in parallel using the target LLM. Notably, Self-Speculative Decoding proposes skipping certain layers to construct the draft model, which eliminates the need for additional parameters or training. Despite its strengths, we observe in this work that drafting with layer skipping exhibits significant sensitivity to domain shifts, leading to a substantial drop in acceleration performance.
%Although this technique has achieved notable speedups, most existing approaches necessitate additional parameters or extensive training to construct effective draft models, thereby restricting their applicability across different LLMs and tasks.
To enhance the domain generalizability of this paradigm, we introduce \method, an algorithm that leverages K-Nearest Neighbor (KNN) search to match different skipped layers with various domain inputs. We evaluated our algorithm in various models and multiple tasks, observing that its application leads to $1.3\times$$\sim$$1.6\times$ speedup in LLM inference.
\footnote{The code is publicly available at: \url{https://github.com/mbsong/KNN-SSD}.}

\end{abstract}

\section{Introduction}

Large language models (LLMs) have proven highly capable in handling various downstream tasks~\citep{touvron2023llama2openfoundation, openai2024gpt4technicalreport, qwen2025qwen25technicalreport}.
However, the token-by-token generation in autoregressive decoding results in quadratic computational complexity, which presents significant efficiency challenges, particularly as the model size increases. 
To address this challenge, speculative decoding (SD) has been proposed as a promising solution for lossless acceleration of LLM inference~\citep{xia-etal-2023-speculative, pmlr-v202-leviathan23a, chen2023acceleratinglargelanguagemodel}. 
At each decoding step, SD uses a lightweight draft model to efficiently predict multiple tokens, which are then verified in parallel by the target LLM to preserve the original output distribution. 
%The effectiveness of SD hinges on the trade-off between drafting latency and speculation accuracy~\citep{xia-etal-2024-unlocking, hu2025speculativedecodingbeyondindepth}. 
%During inference, SD aims to both minimize latency and maximize accuracy to improve efficiency while maintaining output quality.
To enhance inference efficiency while preserving output quality, SD aims to simultaneously minimize latency and maximize accuracy~\citep{xia-etal-2024-unlocking, hu2025speculativedecodingbeyondindepth}. 
Achieving this requires carefully managing the inherent trade-off between the speed of the drafting process and the correctness of the speculative predictions.

\begin{figure}[t]
  \centering
  \includegraphics[width=0.95\columnwidth]{"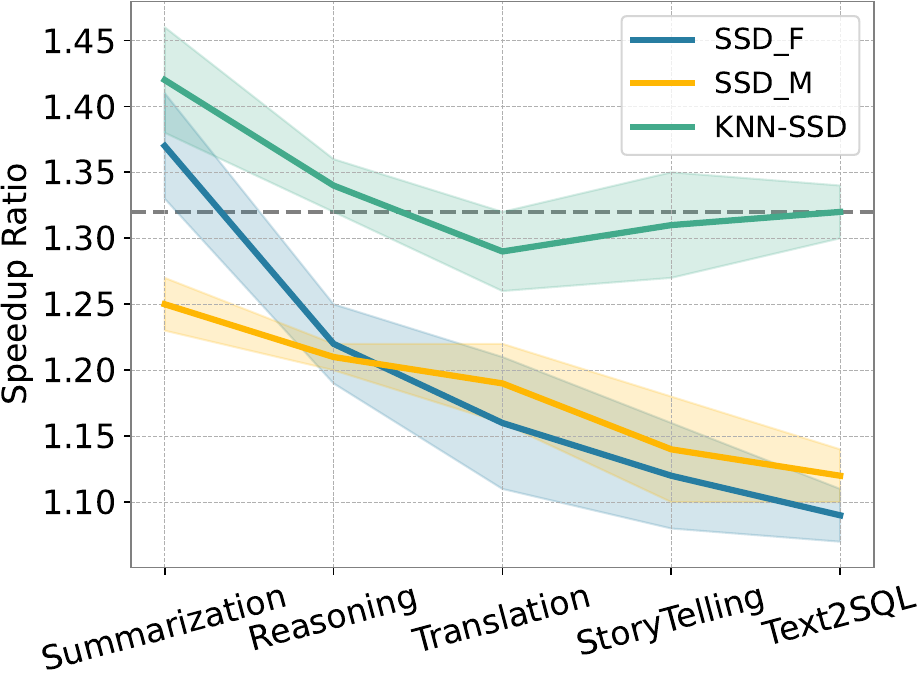"}
  \caption{Average speedup results under task-by-task sample streams. The dashed line represents the average speedup ratio achieved by \method. Results indicate that our \method can achieve a stable speedup while Self-SD methods' speedups decline, as they are sensitive to domain shifts.}
  \label{task_by_task}
\end{figure}

Recent works in SD have yielded significant improvements in the latency-accuracy balance by employing diverse techniques, such as integrating lightweight draft models into LLMs~\citep{ankner2024hydrasequentiallydependentdraftheads, zhang2025learning} or aligning a small model with a larger one~\citep{NEURIPS2023_7b97adea, bachmann2025judgedecodingfasterspeculative} for speculative generation. 
However, these approaches inevitably require additional models, which increase the total number of parameters and introduce additional training complexity. 

%Addressing this concern, Self-SD~\citep{zhang-etal-2024-draft} has been proposed to selectively skip certain layers within the large model itself to construct a compact draft model.
%In this work, we find that \textbf{the selection of skipped layers is not universal}. Instead, one skip-layer configuration could be sensitive to domain shifts.
Self-SD~\citep{zhang-etal-2024-draft} addresses this concern by constructing a compact draft model through the selective skipping of certain layers. 
However, in this work, we find that \textbf{the selection of skipped layers is not universal} and one skip-layer configuration could be sensitive to domain shifts.
For example, when applying a configuration derived from the summarization task to other tasks, as shown in Figure~\ref{task_by_task}, we observe a significant reduction in speedup from 1.35$\times$ to less than 1.10$\times$, highlighting the need for \textit{domain-specific adaptation}. 

To tackle this issue, we propose \method, a method that dynamically adjusts skip-layer configurations based on domain representations. 
The key goal of \method is to \textit{optimize skipped layers specific to each domain, simulate realistic input scenarios, and accurately identify the domain of each sample}.
To achieve this goal, \method integrates three main features: (1) a skipped layer set optimization process for the specific domain of samples, (2) an input sample stream designed to simulate real-life user inputs better, and (3) a KNN model using LLM's last hidden representations to identify the domain of input samples. 
We evaluated our method using LLaMA-2 series~\citep{touvron2023llama2openfoundation} and Qwen-2.5 series~\citep{qwen2025qwen25technicalreport} across various tasks, including summarization, reasoning, translation, storytelling, and text-to-SQL. 
The results indicate that \method achieves a $1.3\times$$\sim$1.6$\times$ speedup compared to autoregressive decoding. 
This approach maintains over $\bf80\%$ token acceptance rate across the LLaMA-2 series and over $\bf99\%$ token acceptance rate across the Qwen-2 series, indicating high alignment potential between the draft model and the target LLM.
Further analysis validated the effectiveness of \method across out-of-domain sample inputs and one dataset that contains various types of samples.

%Figure \ref{task_by_task} and \ref{mean_token_rate} shows results of \method and two Self-Speculative Decoding algorithms on diverse tasks. 
%We designed a scenario where input samples are from one domain to another. 
%As the domain shifts, other methods exhibits significant decline in speedup ratio, mean accepted tokens, and mean acceptance rate. 
%In contrast, \method maintains stable performance across all metrics.

To summarize, our key contributions are:
\begin{enumerate}
    \item We introduce \method, a self-speculative decoding algorithm with a fine-grained skipped layer set selection, which adopts k-nearest neighbor search to retrieve a suitable skipped layer set for each input sample; 
    \item To evaluate our method, we design a dynamic input data stream that contains samples from diverse domains, and \method can achieve a $1.3\times$$\sim$$1.6\times$ speedup across different models without changing the generated tokens' distribution.
\end{enumerate}

\section{Related Work}

\paragraph{Speculative Decoding (SD).}

Speculative Decoding (SD) aims to accelerate autoregressive text generation in LLMs without compromising output quality~\citep{xia-etal-2023-speculative, pmlr-v202-leviathan23a}. 
It reduces decoding latency by predicting multiple future tokens using a draft model or internal mechanisms, followed by verification and correction by the target LLM. 
Existing strategies include aligning small draft models with large models~\citep{xia-etal-2023-speculative, NEURIPS2023_7b97adea, bachmann2025judgedecodingfasterspeculative}, using a retrieval-based model from a datastore~\citep{he2024restretrievalbasedspeculativedecoding, gritta2025dresddenseretrievalspeculative} or predicting $k$ tokens in parallel~\citep{cai2024medusasimplellminference, NEURIPS2024_a79054a9}. 
In another line of work, plug-and-play methods have been examined, with examples including appending pseudo tokens~\citep{fu2024break} and skipping layers dynamically~\citep{metel2024draftflyadaptiveselfspeculative, xia2025swift} during inference. 
Despite efficiency improvement, these methods often rely on auxiliary models or sub-optimal choices, hindering scalability and effectiveness. 
The most related methods to our work include Self-SD~\citep{zhang-etal-2024-draft} and LayerSkip~\citep{elhoushi-etal-2024-layerskip}, which also construct draft models by skipping intermediate LLM layers.
However, both approaches are trained on a single data type and struggle with diverse data streams.
Our work aims to tackle this problem by integrating samples from various domains.

\paragraph{Sparsity and Model Compression.}
Sparsity and model compression are essential for enhancing the efficiency of LLMs by reducing active parameters or computations during inference~\citep{hu2022lora}. 
Common approaches include parameter pruning~\citep{pmlr-v202-frantar23a, ashkboos2024slicegptcompresslargelanguage, sun2024simpleeffectivepruningapproach}, knowledge distillation~\citep{huang2022incontextlearningdistillationtransferring, gu2024minillmknowledgedistillationlarge, wu-etal-2024-lamini}, and quantization~\citep{NEURIPS2022_adf7fa39, liu2023llmqatdatafreequantizationaware, park2024lutgemmquantizedmatrixmultiplication}, which compress models while preserving performance. 
Structured sparsity methods, such as layer skipping~\citep{liu2024acceleratinginferencelargelanguage, bhendawade2024speculativestreamingfastllm, xia2025swift} and dynamic sparsification, further enhance efficiency by adapting computation to input characteristics. 
While these works aim to optimize computational workloads, they may sacrifice performance by using sub-optimal choices because of insufficient search in the layer space. 
In contrast, our \method method can always find optimal choices to accelerate LLM inference losslessly.
% These techniques align closely with speculative decoding strategies, as both aim to optimize computational workloads while preserving model accuracy. 
% Integrating sparsity-aware mechanisms with SD offers a potential avenue for further reducing inference costs in large-scale language models.

\section{Background}

\subsection{Self-Speculative Decoding} % 3 paragraph

Unlike traditional SD methods that require an auxiliary draft model,
% accelerating inference but incurring additional parameters and training complexity.
Self-Speculative Decoding (Self-SD) leverages the LLM's internal structure to draft tokens by selectively skipping certain layers~\citep{zhang-etal-2024-draft}.
Given data $x_1, \dots, x_n$ and the target LLM $\mathbf{M}$ with $L$ layers including both attention and MLP layers, Self-SD aims to find an optimal $z\in\{0,1\}^L$, where $z^{(i)}=1$ indicates that the $i$th layer needs to be skipped and vice versa.
A black-box function $f(\cdot)$ is used to assess the average inference time per verified token:

\begin{equation}
    \label{selfsd}
    z^* = \arg\min_{z} f \big(\mathbf{M}(z)| x_1, \dots, x_n \big).
\end{equation}
Self-SD applies Bayesian optimization~\citep{jones1998efficient} to identify an optimal skip layer set by iteratively selecting new $z$ based on a Gaussian process and evaluating with Eq.\ref{selfsd}. 
After a specified number of iterations, the best $z$ is considered an approximation of $z^*$ and is fixed for inference.
During decoding, the selected layers are skipped to efficiently generate draft tokens, which are then validated in parallel by the full-parameter LLM to ensure the output distribution remains unchanged.

\subsection{Preliminary Study} 
\begin{figure}[t]
  \includegraphics[width=0.95\columnwidth]{"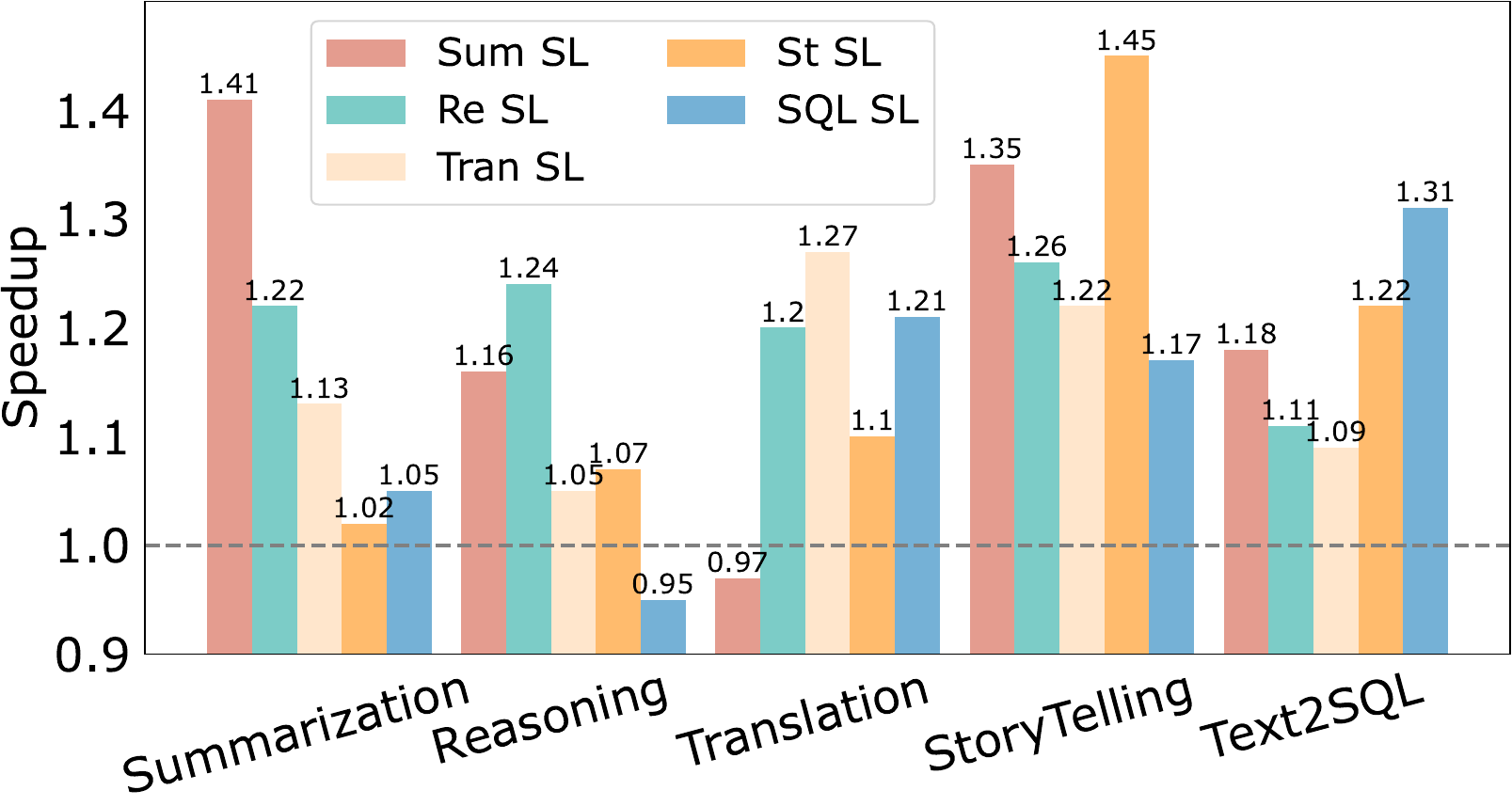"}
  \caption{Different tasks have different optimal skip layer sets. "Sum SL" denotes the skip layer set optimized for the Summarization task. }
  %similarly for the other tasks
  \label{task_specific}
\end{figure}

% A key finding of our study is that the optimal skip-layer configuration varies significantly across different tasks. 
While Self-SD improves inference efficiency, the optimal layers to skip vary significantly across different tasks.
% The optimal layers to skip vary significantly across different tasks, meaning a configuration that works well for one task (e.g., summarization) may degrade performance in another (e.g., mathematical reasoning). 
% This lack of generalizability limits the effectiveness of a static skip-layer configurations, especially in real-world scenarios where query types are unpredictable. 
% To address this, we introduce KNN-SSD, which dynamically selects the best skip-layer configuration based on task characteristics. 
To demonstrate this, we analyze the performance of SD across multiple representative tasks, including summarization, reasoning, storytelling, translation, and text-to-SQL.  As shown in Figure~\ref{task_specific}, an optimized skip-layer configuration for one task does not generalize well to others. 
For example, a configuration that accelerates summarization degrades performance in reasoning tasks.

These results show that the static skip-layer configuration is suboptimal. This limits its effectiveness, particularly in real-world scenarios where query types are unpredictable. 
To achieve both high inference efficiency and minimal performance degradation, task-specific configurations are essential. 
This motivates the development of \method, which dynamically selects the most suitable skip-layer configuration based on task characteristics, ensuring robust and efficient speculative decoding across diverse tasks.

\begin{figure*}[t]
  \includegraphics[width=\textwidth]{"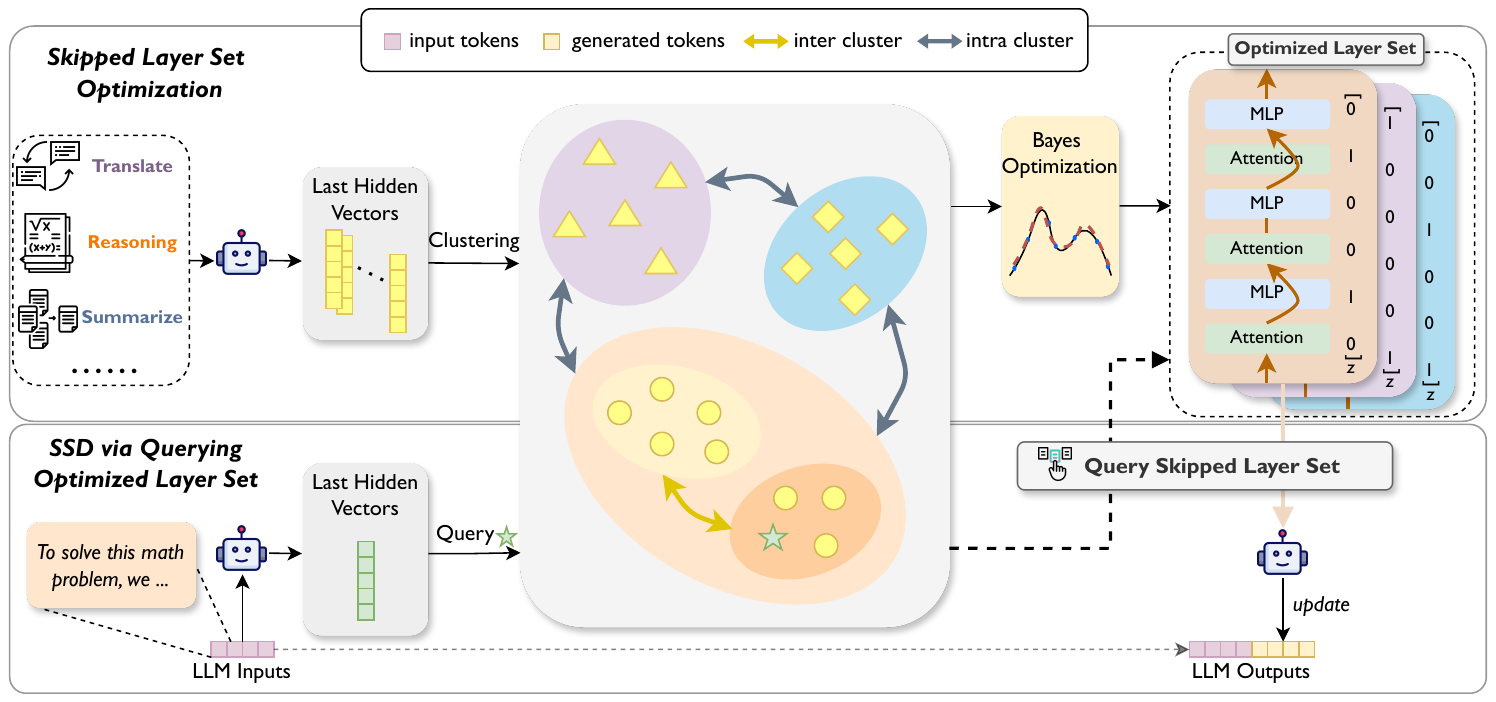"}
  \caption{Layer skipping and KNN process in \method. Before LLM-related generation, \method first performs \textbf{(a) Layer set Searching Optimization.} 
  For each task, \method generates a task-specific skip layer set and stores it in a configuration file; \textbf{(b) Generate Anchor representatives.} 
  \method then produces last hidden vectors for each task to fit a KNN model. 
  When a new sample is input, \method first uses its last hidden vector as the \textbf{input representative} and queries the KNN model. Based on the retrieved result, it selects the corresponding skip layer set to perform decoding, thereby achieving acceleration.}
  \label{main_fig}
\end{figure*}

\section{Methodology}

% explain main figure

% (a) training (knn selection)
% 2 paragraph (clustering -> optimization)
% different domain, use training data for optimization, get representative anchors from each domain (cluster)
% bayes optimization

% (b) inference
% sample -> which cluster (knn) -> optimal layer set
% standard self SD (simplify)

We introduce \method, a generalizable Self-SD method designed to improve inference efficiency while maintaining adaptability across diverse tasks. 
Figure~\ref{main_fig} shows our method of accelerating inference.
It first generates enough last hidden vectors for each task during the pre-inference process. 
Then, a fixed number of vectors are selected as representative anchors to fit a KNN model.
For each task, its optimal skip layer set is searched using a Bayesian optimization process. 
In the inference process, a new input data will find its cluster using the previous KNN model, and the corresponding skip layer set will be used for the target LLM.
Finally, we perform the standard Self-SD process, which contains two stages of drafting and verification to accelerate inference.
By integrating these two processes, \method provides a flexible and effective solution to accelerate LLM inference in real-world applications.

\subsection{Pre Inference}

Given a set of domains $D_1, \dots, D_n$, we first randomly sample multiple instances from each domain, denoted as $d_{i1}, \dots, d_{im}$ for domain $D_i$. 
Each sampled instance $d_{ij}$ is then passed through a pre-trained LLM $\mathbf{M}$ to obtain its last hidden vector representation $v_{ij}$.
These samples are then aggregated and clustered into $n$ groups $\mu_1, \dots, \mu_n$ using the K-means algorithm, where the number of clusters is set to match the number of domains. 
For each cluster $\mu_i$, we identify $k$ representative anchors based on their distance to the cluster centroid.
The collection of selected anchors for cluster $\mu_i$ is denoted as $A_i=\{a_{i1},\dots, a_{ik}\}$, which will be used to fit a KNN model.
The construction of the anchor set $A_i$ is formally defined as follows:

\begin{equation}
    \label{eq:rep}
    A_i = \mathop{\mathrm{arg\,min}}\limits_{S \subseteq D_i, |S| = k} \sum_{v_{ij} \in S} \|v_{ij} - \mu_i\|.
\end{equation}

Subsequently, for each domain $D_i$, we utilize the anchor set $\{a_{i1},\dots,a_{ik}\}$ to determine a domain-specific skip layer set $z_i \in \{0,1\}^L$, where $L$ denotes the total number of layers in the language model $\mathbf{M}$.
Each element $z_i^{(j)}$ indicates whether the $j$-th layer should be skipped $(z_i^{(j)}=1)$ or retained $(z_i^{(j)}=0)$ during inference.
To identify the optimal configuration $z_i$, we employ Bayesian Optimization~\citep{jones1998efficient} over the space of binary layer masks, aiming to minimize an objective black-box function $f(\cdot)$ that measures the average inference time per verified token:

\begin{equation}
    \label{bayes}
    z_i^* = \arg\min_{z_i} f \big(\mathbf{M}(z_i)| a_{i1}, \dots, a_{ik} \big).
\end{equation}
All $z_1^*, \dots, z_n^*$ will be stored for future use and not be changed.

\begin{table}[ht]
  \centering
  \small
  \begin{tabular}{@{}llcc@{}}
    \toprule
    \multirow{2}{*}{\textbf{Models}} & \multirow{2}{*}{\textbf{Methods}}
     & \multirow{2}{*}{\makecell{\textbf{Speed} \\ \textbf{(token/s)}}} & 
     \multirow{2}{*}{\makecell{\textbf{Overall} \\ $\mathbb{E}$(Spd.)}} \\

     & & & \\
     
    \midrule

    \multirow{4}{*}{LLaMA-2-13B} 
    & \texttt{Vanilla}  & 13.62 & 1.00$\times$ \\
    & \ssdf  & 16.34 & 1.20$\times$ \\
    & \ssdm  & 16.88 & 1.24$\times$ \\
    & \method  & \textbf{19.61} & \textbf{1.44$\times$} \\

    \midrule

    \multirow{4}{*}{\makecell{LLaMA-2-13B \\ -Chat}} 
    & \texttt{Vanilla}  & 13.22 & 1.00$\times$ \\
    & \ssdf  & 14.67 & 1.11$\times$ \\
    & \ssdm  & 15.33 & 1.16$\times$ \\
    & \method  & \textbf{17.85} & \textbf{1.35$\times$} \\

    \midrule

    \multirow{4}{*}{Qwen-2.5-14B} 
    & \texttt{Vanilla} & 11.16 & 1.00$\times$ \\
    & \ssdf  & 14.06 & 1.26$\times$ \\
    & \ssdm  & 15.40 & 1.38$\times$ \\
    & \method & \textbf{18.08} & \textbf{1.62$\times$} \\

    \midrule

    \multirow{4}{*}{\makecell{Qwen-2.5-14B\\-Instruct}} 
    & \texttt{Vanilla} & 10.79 & 1.00$\times$ \\
    & \ssdf  & 12.84 & 1.19$\times$ \\
    & \ssdm  & 13.49 & 1.25$\times$ \\
    & \method & \textbf{16.30} & \textbf{1.51$\times$} \\
    
    \bottomrule
  \end{tabular}  
  \caption{Comparison between \method and two Self-SD methods. We report average decoding speed (token/s) under greedy decoding and average speedup ratio among four different mix ratios. More details are provided in Appendix~\ref{result_detail}.}
  \label{main}
\end{table}

\subsection{Inference}

For a newly arrived sample $s$, we first extract its last hidden vector $v$ from the model. 
We then perform a KNN search based on cosine similarity between the hidden vector of $s$ and all representative anchors. 
This process yields a corresponding domain label, effectively classifying the sample into one of the known domains $i^*$. 
Based on the identified domain, we apply its associated optimal skip-layer configuration $z_{i^*}^*$ to $\mathbf{M}$ to accelerate inference:

\begin{align}
    \label{eq:knn}
        i^*, j^* = \arg&\max_{i,j} \frac{v \cdot a_{ij}}{\|v\| \cdot \|a_{ij}\|}, \\
        \text{Domain}&(s) = i^*,\\
        \mathbf{M} &\leftarrow z_{i^*}^*.
\end{align}

We then perform the standard Self-SD process~\citep{zhang-etal-2024-draft}, which involves two stages: drafting and verification. 
During the drafting stage, the LLM uses the previously selected skip-layer configuration $z_i$ as a draft model $M(z_i)$ to generate a sequence of draft tokens:
\begin{equation}
    y' = \arg\max_{y} \text{log} \ P(y \ |\ x,\ \mathbf{y};\ M(z_i)),
\end{equation}
where $x$ and $\mathbf{y}$ denote input and output generated by LLM, respectively, and $y'$ represents the token produced by the autoregressive process.
In the verification stage, the full LLM verifies the draft tokens in a single forward pass. This step validates the correctness of the generated tokens and either accepts them or triggers a re-drafting if discrepancies are found.

To better simulate real-world task streams, we introduce the mix ratio $r$, which denotes the probability that the next input sample belongs to a different task than the current one. 
A mix ratio of 0 corresponds to a task-by-task input stream, where all consecutive samples come from the same task. In contrast, a mix ratio of 1 indicates maximum task mixing, where every two consecutive samples are from different tasks. As the mix ratio grows, the frequency of domain shift increases.

\begin{equation}
\label{eq:randomness}
    \scalebox{0.9}{$
        P(s_{i+1} \in D_j \mid s_i \in D_k) =
            \begin{cases}
            \frac{r}{N-1} & \text{if } j \neq k, \\
            1 - r & \text{if } j = k.
            \end{cases}
    $}
\end{equation}

\section{Experiments}

\subsection{Experimental Setup}

\paragraph{Implementation Details.}
We mainly evaluate \method on LLaMA-2 series~\citep{touvron2023llama2openfoundation} and Qwen-2.5 series~\citep{qwen2025qwen25technicalreport} across various tasks, including summarization, mathematical reasoning, storytelling, translation, and text-to-SQL. 
The evaluation datasets include CNN/Daily Mail (CNN/DM)~\citep{nallapati-etal-2016-abstractive}, GSM8K~\citep{gsm8k}, TinyStories~\citep{tinystories}, Wmt16 DE-EN (Wmt16)~\citep{bojar-EtAl:2016:WMT1}, and Spider2~\citep{lei2025spider20evaluatinglanguage}, where more details can be found in Appendix~\ref{datasets}.

For each dataset, we used Bayesian optimization \footnote{\url{https://github.com/bayesian-optimization/BayesianOptimization}} (BO) to perform 1,000 iterations in 8 representative samples in search of the optimal skip-layer configuration.
As detailed in Appendix~\ref{pre}, the optimal layer configurations are task-specific.
The representative samples are selected via the K-means algorithm from all last hidden vectors generated by the LLM in the corresponding dataset, ensuring optimal coverage of the feature space.
The maximum generation lengths on CNN/DM, GSM8K, Wmt16, Spider2, and TinyStories are set to 128, 128, 128, 128, and 512, respectively. 
We conduct 1-shot evaluation for CNN/DM and TinyStories, 3-shot evaluation for Spider2, and 5-shot evaluation for GSM8K and Wmt16.
For each dataset, we extracted the most representative $k=10$ hidden vectors from the last hidden layer across all data samples using cosine similarity to serve as anchor points for the KNN model, following the same approach as introduced earlier in the BO framework.
For each new input sample, we also compute the cosine similarity between its last hidden vector and the anchors, and assign it to the task of its nearest neighbor.
Detailed setups are presented in Appendix~\ref{setup}.

\paragraph{Baselines.}
In our primary experiments, we compared \method and Self-SD approach~\citep{zhang-etal-2024-draft} to assess their effectiveness.
For the Self-SD method, we primarily simulated two scenarios. 
In the first scenario, a fixed skip-layer configuration was determined based on the first sample in the task stream and remained unchanged throughout the process, which is denoted as \ssdf.
In the second scenario, the skip-layer configuration was adjusted by re-performing BO according to the task distribution within the stream, and the newly searched configuration was subsequently applied for inference and also remained unchanged, which is denoted as \ssdm.

\begin{table}[t]
    \centering
    \small
    \setlength{\tabcolsep}{3pt}
    \begin{tabular}{@{}llcccc@{}}
        \toprule
        \multirow{2}{*}{\textbf{Models}} & \multirow{2}{*}{\textbf{Methods}}
        & $r$=0.0 & $r$=0.3 & $r$=0.7 & $r$=1.0 \\
     
        & & $\mathbb{E}$(Spd.) & $\mathbb{E}$(Spd.) & $\mathbb{E}$(Spd.) & $\mathbb{E}$(Spd.) \\
        
        \midrule
        \multirow{4}{*}{\makecell{LLaMA\\-2-13B}} 
        & \texttt{Vanilla} & 1.00$\times$ & 1.00$\times$ & 1.00$\times$ & 1.00$\times$ \\
        & \ssdf & 1.24$\times$  & 1.21$\times$ & 1.19$\times$ & 1.17$\times$ \\
        & \ssdm & 1.23$\times$  & 1.27$\times$ & 1.24$\times$ & 1.23$\times$ \\
        & \method & \textbf{1.42$\times$}  & \textbf{1.45$\times$}  & \textbf{1.43$\times$} & \textbf{1.45$\times$} \\
        
        \midrule
        \multirow{4}{*}{\makecell{Qwen\\-2.5-14B}} 
        & \texttt{Vanilla} & 1.00$\times$ & 1.00$\times$ & 1.00$\times$ & 1.00$\times$ \\
        & \ssdf & 1.25$\times$  & 1.23$\times$ & 1.27$\times$ & 1.28$\times$ \\
        & \ssdm & 1.40$\times$  & 1.36$\times$ & 1.39$\times$ & 1.38$\times$ \\
        & \method & \textbf{1.60$\times$}  & \textbf{1.64$\times$}  & \textbf{1.63$\times$} & \textbf{1.61$\times$} \\
        
        \bottomrule
    \end{tabular}
    \caption{Our experiments across four mix ratio settings (0.0, 0.3, 0.7, and 1.0) reveal that \ssdf performs well only in the early stages, but suffers degradation as the input distribution changes. \ssdm, which applies a unified sub-optimal configuration, performs consistently worse. In contrast, \method maintains stable speedup across all mix ratios, validating its robustness to distribution shifts.}
    \label{diff_mix_ratio}
\end{table}

\paragraph{Evaluation Metrics.}
We evaluate \method using two standard metrics commonly adopted in evaluation: the mean generated length $M$~\citep{NEURIPS2018_c4127b91} and the token acceptance rate $\alpha$~\citep{pmlr-v202-leviathan23a}. 
Beyond these, we also report the expected decoding throughput in tokens per second, along with the expected wall-time speedup ratio compared to standard autoregressive decoding. 
Given $M$ and $\alpha$, the expected speedup can be derived by the formula given by~\citet{pmlr-v202-leviathan23a}:
\begin{equation}
    \mathbb{E}\text{(Spd.)} = \frac{M\alpha}{(M-1)(1-s)+\alpha}
\end{equation}
where $s$ denotes the ratio of skipped layers.
More details about this formula are discussed in Appendix~\ref{eval_metrics}.

\begin{table}[t]
    \centering
    \small
    \setlength{\tabcolsep}{4pt}
    \begin{tabular}{@{}llccc@{}}
        \toprule
        \textbf{Models} &\textbf{Methods} & $M$ & $\alpha$ & \textbf{Speedup}  \\
        \midrule
        \multirow{4}{*}{\makecell{LLaMA-2\\-13B}} 
        & \texttt{Vanilla} & 1.00 & - & 1.00$\times$ \\
        & \ssdf & 2.17 & 0.62 & 1.10$\times$ \\
        & \ssdm & 2.53 & 0.68 & 1.14$\times$ \\
        & \method & \textbf{3.12} & \textbf{0.88} & \textbf{1.34$\times$} \\
        \midrule
        \multirow{4}{*}{\makecell{LLaMA-2\\-13B-Chat}} 
        & \texttt{Vanilla} & 1.00 & - & 1.00$\times$ \\
        & \ssdf & 1.97 & 0.57 & 1.04$\times$ \\
        & \ssdm & 2.14 & 0.59 & 1.09$\times$ \\
        & \method & \textbf{2.87} & \textbf{0.85} & \textbf{1.28$\times$} \\
        \bottomrule
    \end{tabular}
    \caption{The results demonstrate the mean accepted tokens, token acceptance rate, and actual speedup ratio obtained from our tests on the LLaMA-2 series, conducted under a mix ratio of 1.0 across the five main datasets (CNN/DM, GSM8K, Wmt16, Spider2, and TinyStories), showing that \method outperforms two Self-SD methods in every metric.}
    \label{real_speedup}
\end{table}

\begin{figure}[t]
  \centering
  \includegraphics[width=0.9\columnwidth]{"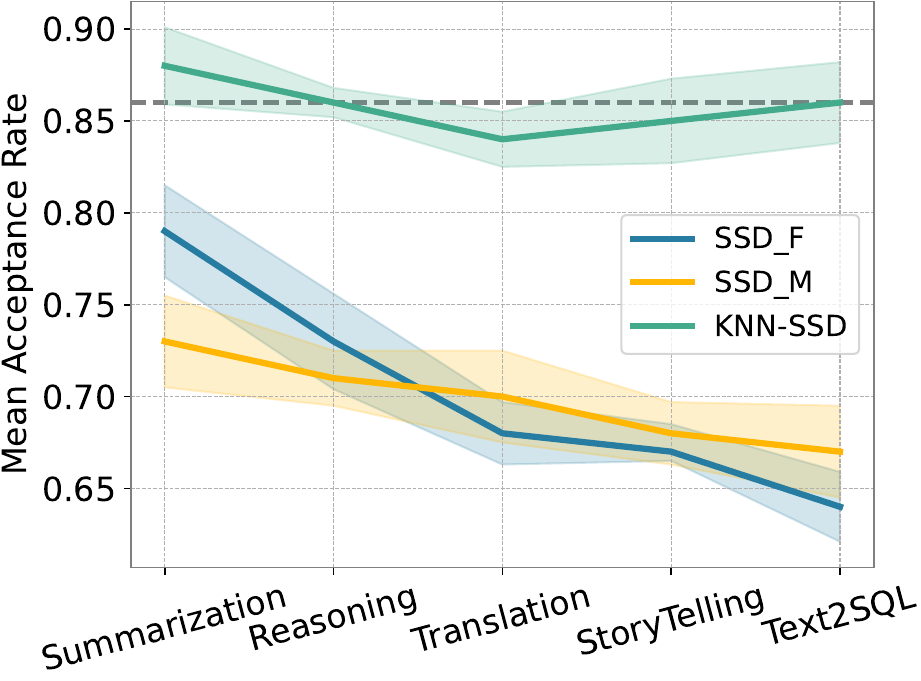"}
  \includegraphics[width=0.9\columnwidth]{"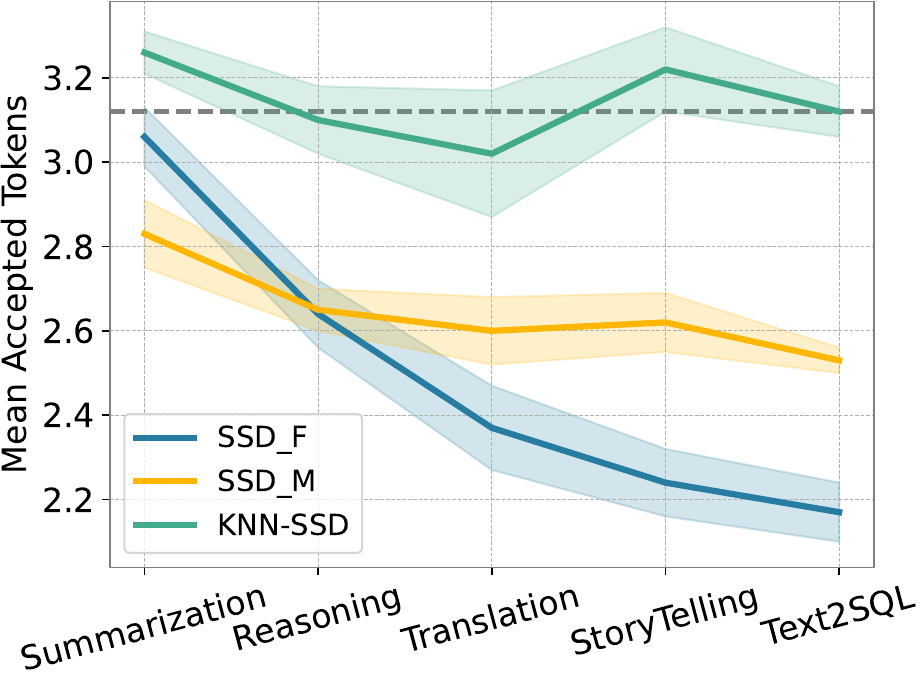"}
  \caption{The mean accepted tokens and mean acceptance rate under task-by-task sample streams. The dashed lines represent the average length and rate achieved by \method across all five datasets.}
  \label{mean_token_rate}
\end{figure}

\subsection{Main Result}

%Table~\ref{main} presents the comparison between \method and two Self-SD methods on generation tasks.
We compare \method with two Self-SD methods on generation tasks: Table~\ref{main} reports the average speedup, while Table~\ref{diff_mix_ratio} details the results across different configurations.
In our experiments, we evaluate \method under four settings: mix ratio = 0, 0.3, 0.7, and 1 separately, with 40 samples from five datasets each, 200 samples in total.
The experimental results demonstrate the following findings: 
(1) \method shows superior efficiency over prior methods, achieving consistent speedups of $1.35\times$$\sim$$1.62\times$ over vanilla autoregressive decoding across various models. 
(2) The mix ratio of sample flows doesn't affect the speedup of \method.
%As shown in Table~\ref{main}, \method produces a mean generated length $M$ of 3.23. 
The speedup remains stable, which indicates that \method can handle various samples in a more realistic scenario.

We present the mean accepted tokens, acceptance rate, as well as the actual speedup of LLaMA-2-13B series in Table~\ref{real_speedup} and Figure~\ref{mean_token_rate}, which further validates the superiority of \method over Self-SD.

\begin{figure}[t]
  \centering
  \includegraphics[width=0.9\columnwidth]{"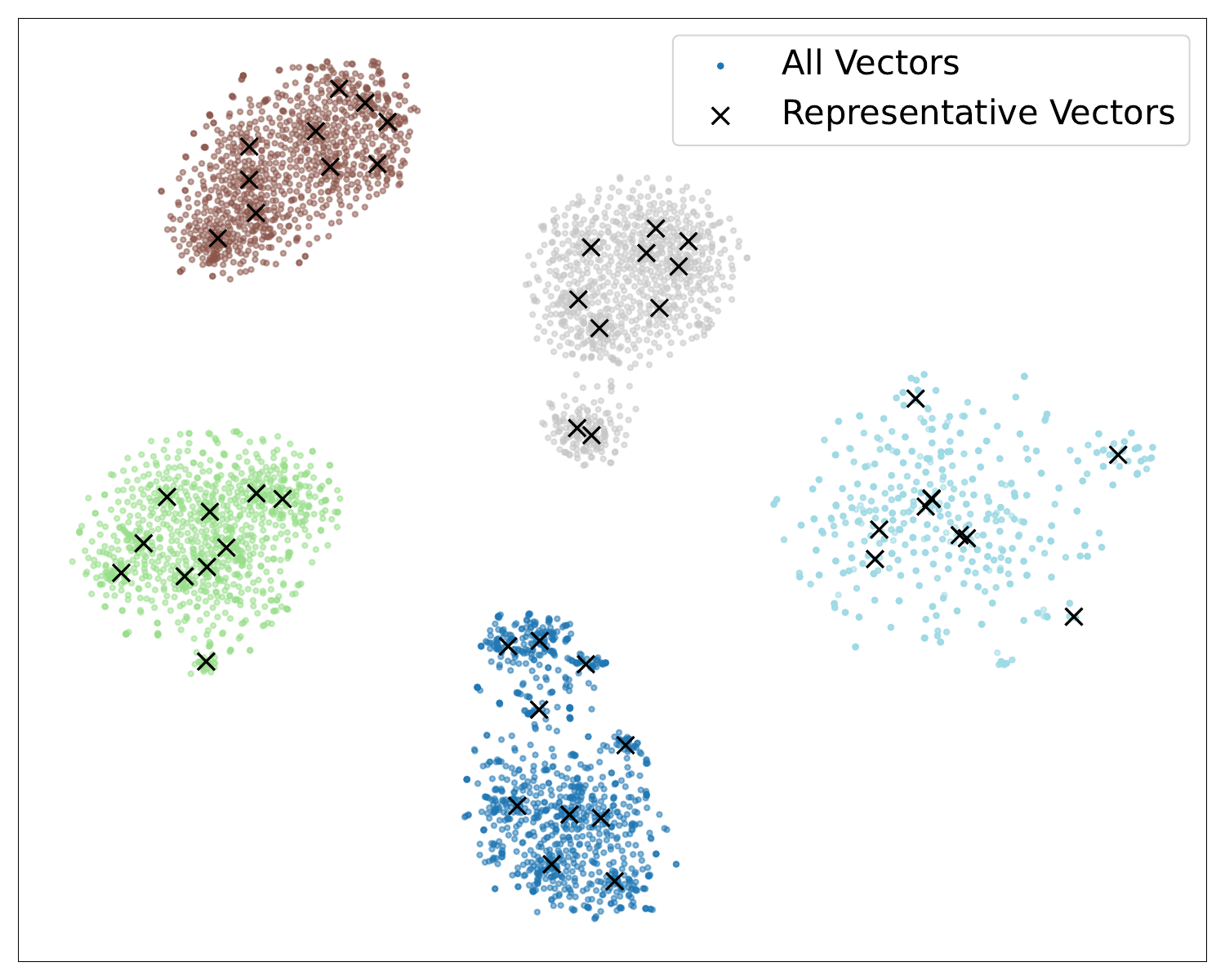"}
  \caption{Visualization of last hidden vectors from five domains of samples. Results show these vectors can be clearly divided into five clusters. From each cluster, we selected ten vectors as representative anchors for our KNN model.}
  \label{KNN}
\end{figure}

\subsection{Analysis}

\paragraph{Pre-inference Overhead.}
The only additional overhead introduced by \method is the storage of the hidden states and the execution of the KNN search, which is negligible.
Specifically, when the maximum decoding length is set to 128, the average generation latency across the five datasets on LLaMA-2-13B is approximately 7.5 seconds, of which storing the last hidden state and conducting the KNN search takes around 0.18 seconds, accounting for only about 2.5\% of the total.
Under longer decoding lengths such as 512 or 1024 tokens, the average generation time increases to about 23.2 seconds or 43.7 seconds, while the KNN search time remains roughly constant, resulting in a proportion of just 0.77\% or 0.41\%.
Meanwhile, the classification accuracy for all datasets is 100\%, as shown in Figure~\ref{KNN}. This indicates that our KNN-based domain partitioning yielded perfectly accurate clustering in these cases.

\begin{figure}[t]
  \centering
  \includegraphics[width=0.9\columnwidth]{"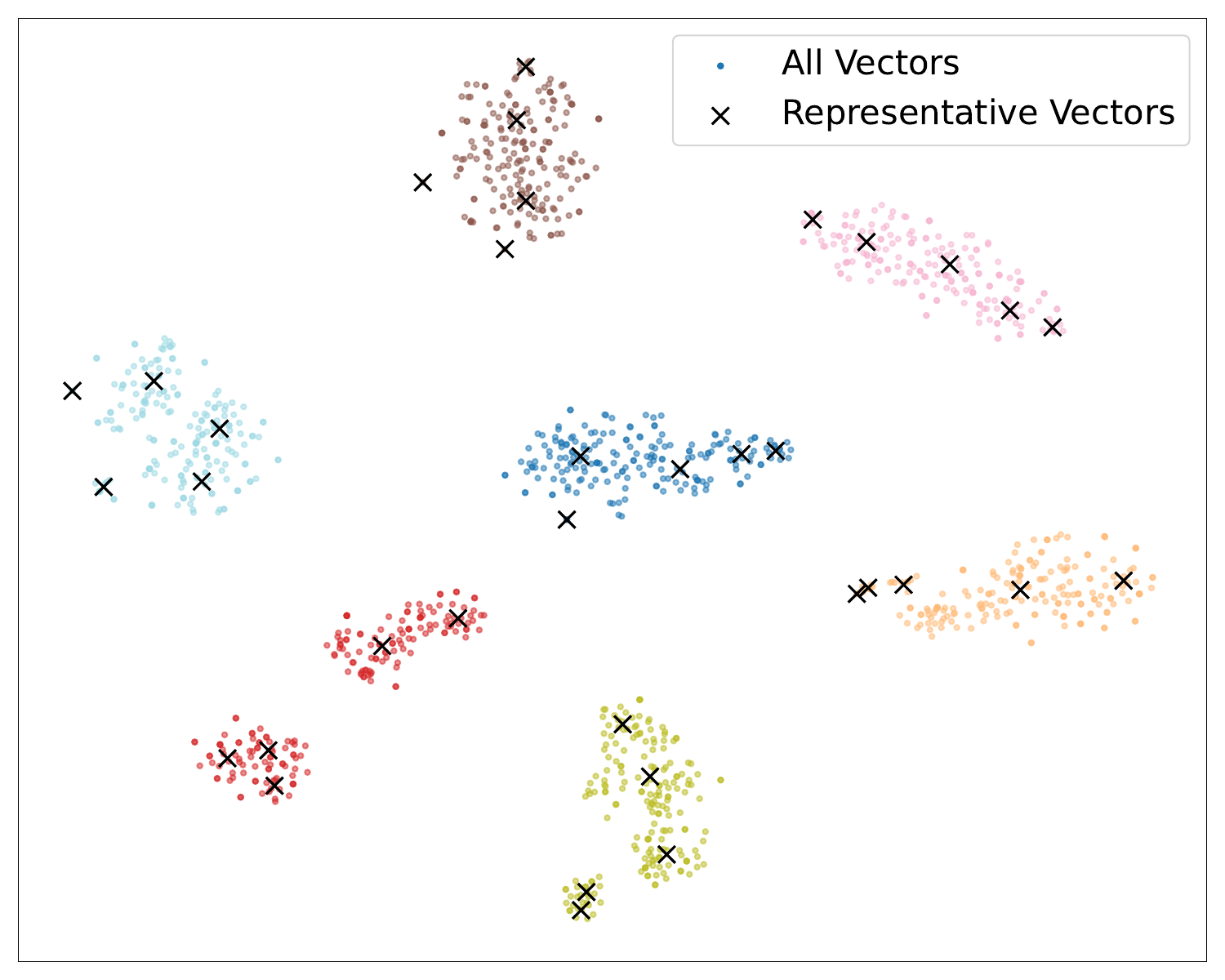"}
  \caption{Visualization of 700 last hidden vectors from the MATH dataset using t-SNE method. It is clear to see that all vectors can be categorized into 7 groups, which aligns with the fact that the MATH dataset has 7 different kinds of problems.}
  \label{math}
\end{figure}

\paragraph{Inter \& Intra.}
We use the MATH~\citep{hendrycks2021measuringmathematicalproblemsolving} dataset to assess the capabilities of \method in a single dataset with multiple domains.
In the MATH dataset, math questions are categorized into seven types. 
Thus, using one specific skip layer set for this dataset is insufficient, and we introduce a fine-grained clustering to handle this mixed domain. 
Figure~\ref{math} shows that each type of math question can be clustered into a single group.
Table~\ref{math_result} indicates the speedup result for each method, where we can clearly see that \method outperforms Self-SD methods and achieves a speedup of 1.23$\times$ and a mean generated length $M$ of 2.37.

\begin{table}[t]
    \centering
    \begin{tabular}{@{}lccc@{}}
        \toprule
        \textbf{Methods} & $M$ & $\alpha$ & \textbf{Speedup} \\
        \midrule
        \texttt{Vanilla} & 1.00 & - & 1.00$\times$ \\
        \ssdf & 1.54 & 0.51 & 0.97$\times$ \\
        \ssdm & 1.82 & 0.59 & 1.02$\times$ \\
        \method & \textbf{2.37} & \textbf{0.81} & \textbf{1.23$\times$} \\
        \bottomrule
    \end{tabular}
    \caption{Results of MATH dataset using LLaMA-2-13B.}
    \label{math_result}
\end{table}

\begin{figure}[ht]
  \centering
  \includegraphics[width=0.95\columnwidth]{"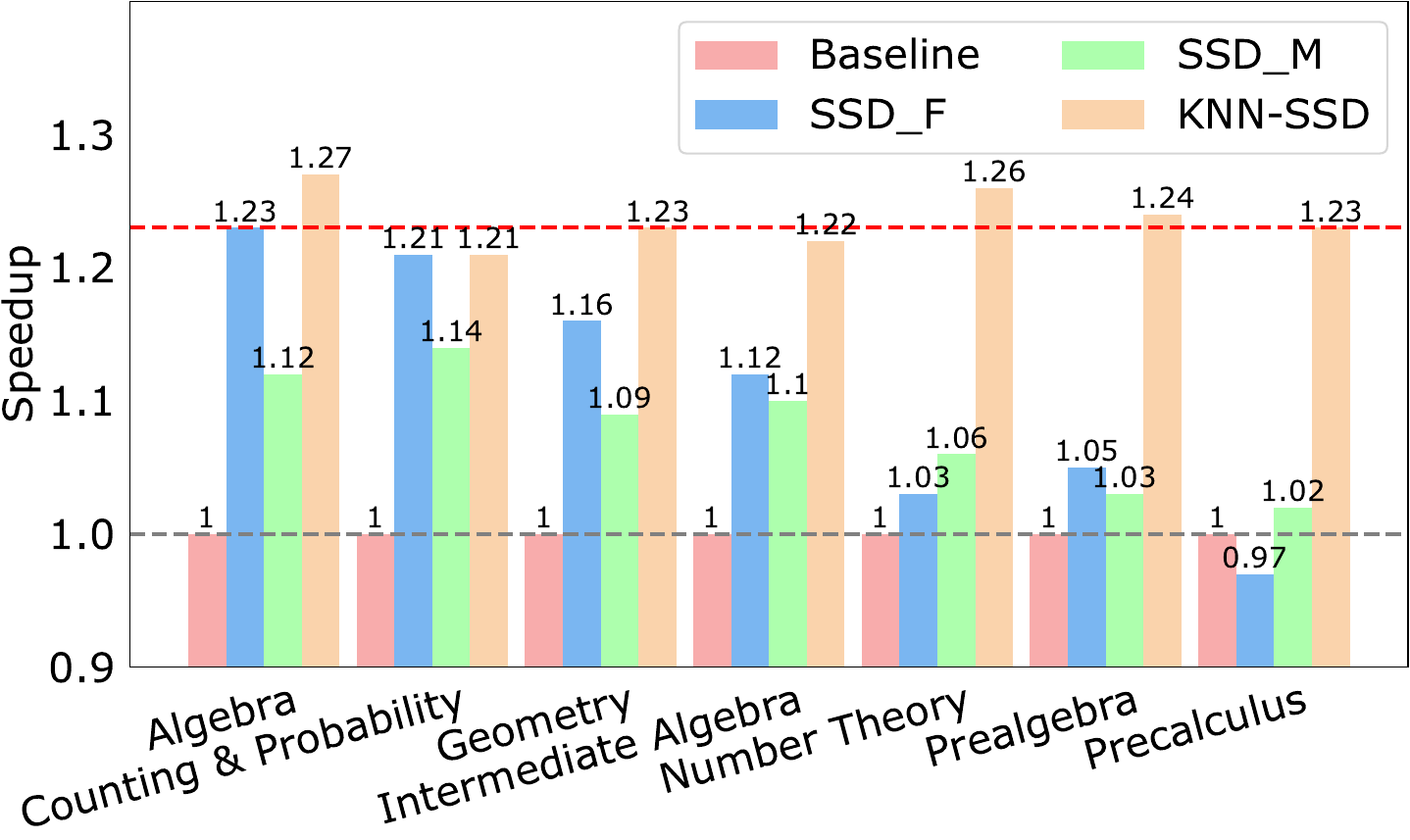"}
  \caption{Speedup results for task-by-task sample streams on MATH dataset under three methods. While \method maintains a speedup of around 1.25$\times$, two Self-SD methods decline as the number of domains grows.}
  \label{math_speedup}
\end{figure}

Figure~\ref{math_speedup} visualizes speedups on task-by-task settings. \ssdf achieves high performance only on the first subtask and declines on the rest of the subtasks; while \method has a better speedup than the other two Self-SD methods.

\paragraph{Out of Domain Generalization.}
We adopt the XSUM~\citep{xsum-emnlp} dataset, the MATH~\citep{hendrycks2021measuringmathematicalproblemsolving} dataset, and the Alpaca~\citep{alpaca} dataset as out-of-domain tasks to assess \method's generalizability. XSUM and CNN/DM datasets belong to summarization tasks, whereas MATH and GSM8K datasets involve reasoning-based tasks. 
Therefore, although we did not search for their respective optimal skip layer sets for XSUM and MATH in our experiments, it is reasonable that \method would assign XSUM samples to CNN/DM and thus adopt CNN/DM’s optimal skip layer set, and the same applies to MATH samples.

\begin{table}[ht]
    \centering
    \small
    \begin{tabular}{@{}llccc@{}}
        \toprule
        \textbf{Datasets} &\textbf{Methods} & $M$ & $\alpha$ & \textbf{Speedup}  \\
        \midrule
        \multirow{3}{*}{\textbf{XSUM}} 
        & \texttt{Vanilla} & 1.00 & - & 1.00$\times$ \\
        & \texttt{Self-SD} & 1.42 & 0.56 & 0.99$\times$ \\
        & \method & \textbf{2.51} & \textbf{0.84} & \textbf{1.24$\times$} \\
        
        \midrule
        \multirow{3}{*}{\textbf{MATH}} 
        & \texttt{Vanilla} & 1.00 & - & 1.00$\times$ \\
        & \texttt{Self-SD} & 1.34 & 0.48 & 0.93$\times$ \\
        & \method & \textbf{2.13} & \textbf{0.76} & \textbf{1.17$\times$} \\
        
        \midrule
        \multirow{3}{*}{\textbf{Alpaca}} 
        & \texttt{Vanilla} & 1.00 & - & 1.00$\times$ \\
        & \texttt{Self-SD} & 1.26 & 0.43 & 0.92$\times$ \\
        & \method & \textbf{1.95} & \textbf{0.67} & \textbf{1.15$\times$} \\
        \bottomrule
    \end{tabular}
    \caption{Results of out-of-domain datasets using LLaMA-2-13B-Chat. No representative anchor of these three domains is generated.}
    \label{out_of_domain}
\end{table}

Compared to these two datasets, the Alpaca dataset contains more diverse instruction-answer pairs across summarization, reasoning, grammar, and many other tasks. Results indicate that although some of the domains are not covered by our five datasets in the main experiments, our method can still assign an unknown sample to its most similar domain and thus achieve inference acceleration. As shown in Table~\ref{out_of_domain}, our experimental results demonstrate that the model achieves an approximately $1.15\times\sim1.25\times$ speedup under the \method method, even without prior search.

\paragraph{Number of Clusters.} Table~\ref{alpaca_result} shows the result of the influence of cluster numbers. 
We conducted experiments on the Alpaca dataset as it covers a variety of domains, using K-means clustering with varying numbers of clusters. 
As shown in the results, the speedup effect improves as the number of clusters increases, eventually surpassing the speedup ratio observed in the out-of-domain experiments (Table~\ref{out_of_domain}). 
However, when the cluster count exceeds 5 (e.g., up to 7), the speedup plateaus, indicating that partitioning Alpaca into five clusters is sufficient—further subdivision yields no additional gains.

\begin{table}[ht]
    \centering
    \small
    \begin{tabular}{cccc}
        \toprule
        \textbf{Num.} & $M$ & $\alpha$ & \textbf{Speedup} \\
        \midrule
        1 & 1.86 & 0.65 & 1.05$\times$ \\
        3 & 2.20 & 0.75 & 1.17$\times$ \\
        5 & 2.52 & 0.80 & 1.23$\times$ \\
        7 & 2.55 & 0.81 & 1.23$\times$ \\
        \bottomrule
    \end{tabular}
    \caption{Results of the Alpaca dataset among different numbers of clusters using LLaMA-2-13B-Chat. \textbf{Num.} denotes the number of clusters.}
    \label{alpaca_result}
\end{table}

\begin{figure}[ht]
    \centering
    \includegraphics[width=0.95\columnwidth]{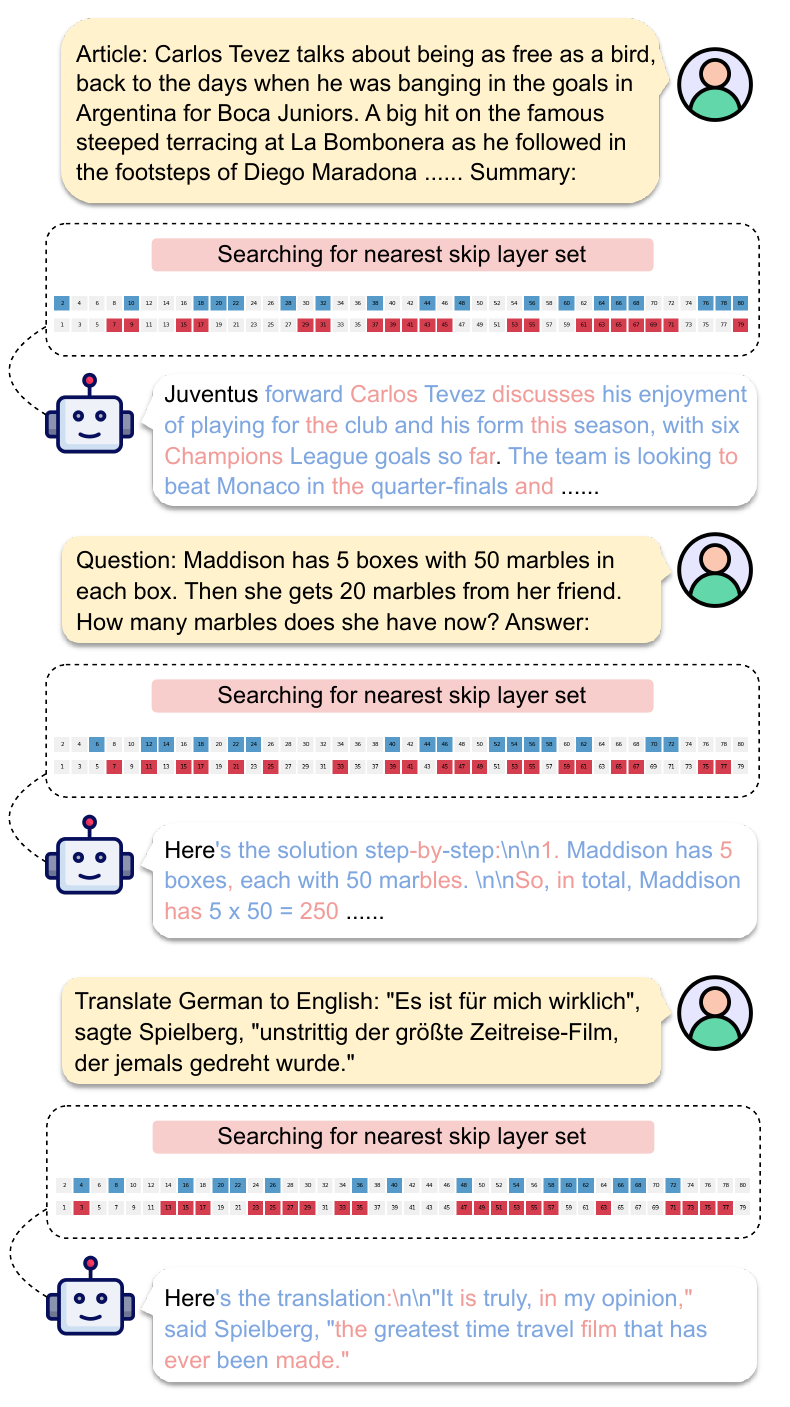}
    \caption{Case study of how \method works. Blue tokens indicate that they are generated during the drafting step and verified by the model, while red tokens indicate they are generated by prediction from the verification step. Squares in red and blue indicate skipped attention layers and MLP layers, respectively.}
    \label{case_study}
\end{figure}

\paragraph{Case Study.} To better illustrate how our method works, we provide a case study that presents a typical sample stream. In Figure~\ref{case_study}, a sample stream contains three common types of queries a user might ask: summarization, reasoning, and translation. For each input query, \method will first compute its last hidden vector and then use a KNN model to find its optimal skipped layer set. The typical speculative decoding will be conducted with draft and verification steps, where blue and red tokens indicate that they are generated separately in draft and verification steps. By constantly changing skipped layer sets, \method achieves a stable speedup compared to other methods that use a static strategy, which is insufficient for diverse inputs.

\section{Conclusion}

In this work, we introduce \method, an algorithm that leverages K-Nearest Neighbor search to match suitable skipped layers for various domain inputs. \method is designed to find an optimal skipped layer set for each domain of data, which accelerates LLM's inference losslessly. To assess its ability, we define a mix ratio of a sample stream, indicating how frequently the domain changes.
We conducted extensive experiments with various LLMs and mix ratios and found that \method can achieve a speedup of around $1.3\times$$\sim$$1.6\times$ without changing the ordinary distribution of the generated tokens. 
Our in-depth analysis indicates that a single dataset may also contain mixed domains. Furthermore, \method can achieve a $1.2\times$ speedup on out-of-domain datasets, showing its great potential in handling various data streams in real-life scenarios.

\section*{Limitations}

A few limitations need to be considered while our \method achieves a notable speedup on various models. 
First, we did not incorporate draft tree verification, which has been shown to improve the token acceptance rate~\citep{xia2025swift}.
Second, our current evaluation is limited to models of moderate scale. 
Due to practical considerations related to computational resources, we have not yet extended our method to larger-scale models. 
We leave these directions for future work.

\section*{Ethics Statement}
\label{subsec:ethics} 
The datasets used in our experiment are publicly released and labeled through interaction with humans in English. In this process, user privacy is protected, and no personal information is contained in the dataset. The scientific artifacts that we used are available for research with permissive licenses. And the use of these artifacts in this paper is consistent with their intended use. Therefore, we believe that our research work meets the ethics of ACL.

\section*{Acknowledgement}
\label{subsec:ack}
We thank the anonymous reviewers for their helpful
comments on this paper. This work was partially
supported by National Natural Science Foundation
of China (No. 62476010).

\bibliography{anthology,custom}

\clearpage
 
\appendix

\section{Preliminary Details}
\label{pre}

\begin{figure*}[ht]
    \centering
    \begin{subfigure}[t]{\textwidth}
        \includegraphics[width=\textwidth]{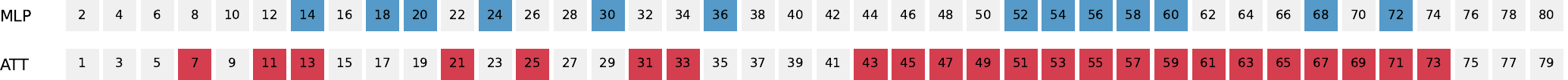}
        \caption{Summarization - CNN/DM}
        \label{llama_cnndm}
    \end{subfigure}%
    \vspace{0.5cm}
    \begin{subfigure}[t]{\textwidth}
        \includegraphics[width=\textwidth]{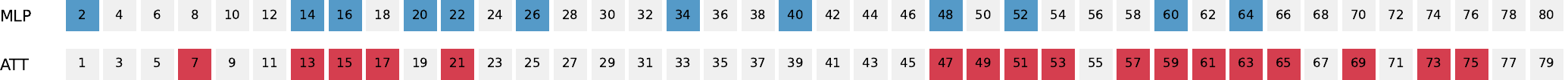}
        \caption{Reasoning - GSM8K}
        \label{llama_gsm8k}
    \end{subfigure}%
    \vspace{0.5cm}
    \begin{subfigure}[t]{\textwidth}
        \includegraphics[width=\textwidth]{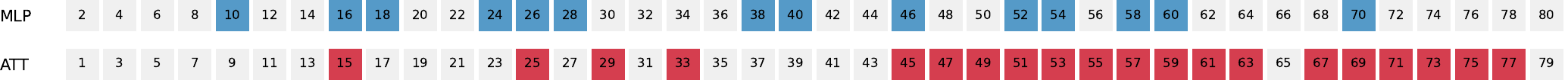}
        \caption{Translation - WMT16}
        \label{llama_wmt16}
    \end{subfigure}%
    \vspace{0.5cm}
    \begin{subfigure}[t]{\textwidth}
        \includegraphics[width=\textwidth]{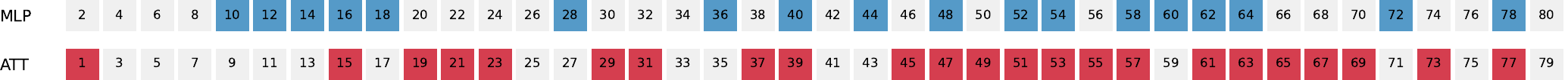}
        \caption{Storytelling - TinyStories}
        \label{llama_story}
    \end{subfigure}%
    \vspace{0.5cm}
    \begin{subfigure}[t]{\textwidth}
        \includegraphics[width=\textwidth]{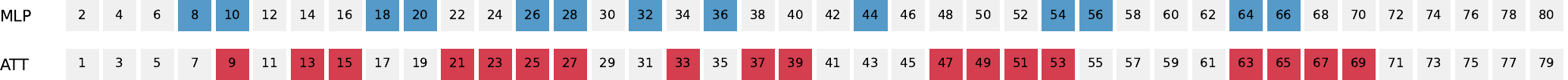}
        \caption{Text-to-SQL - Spider2}
        \label{llama_sql}
    \end{subfigure}
    \caption{Visualization of skipped layer set configuration of LLaMA-2-13B optimized by Self-SD \citep{zhang-etal-2024-draft} on different task domains. Gray squares indicate retained layers, red squares denote skipped attention layers, and blue squares signify skipped MLP layers.}
    \label{llama_skip}
\end{figure*}

\begin{figure*}[ht]
    \centering
    \begin{subfigure}[t]{\textwidth}
        \includegraphics[width=\textwidth]{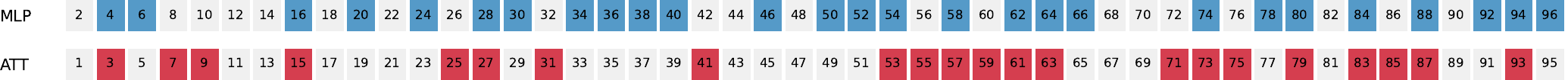}
        \caption{Summarization - CNN/DM}
        \label{qwen_cnndm}
    \end{subfigure}%
    \vspace{0.5cm}
    \begin{subfigure}[t]{\textwidth}
        \includegraphics[width=\textwidth]{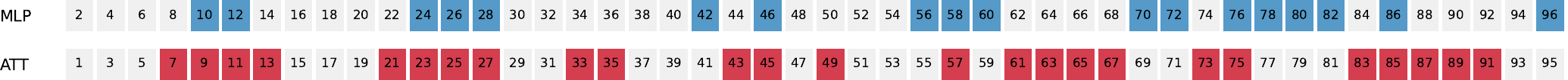}
        \caption{Reasoning - GSM8K}
        \label{qwen_gsm8k}
    \end{subfigure}%
    \vspace{0.5cm}
    \begin{subfigure}[t]{\textwidth}
        \includegraphics[width=\textwidth]{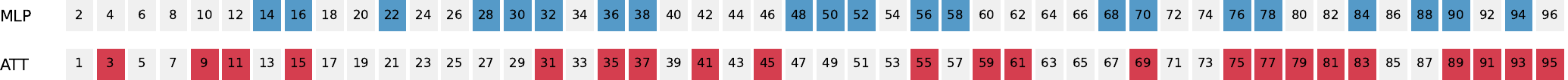}
        \caption{Translation - WMT16}
        \label{qwen_wmt16}
    \end{subfigure}%
    \vspace{0.5cm}
    \begin{subfigure}[t]{\textwidth}
        \includegraphics[width=\textwidth]{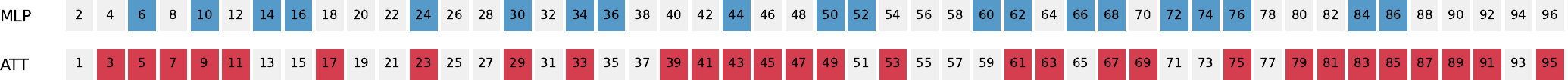}
        \caption{Storytelling - TinyStories}
        \label{qwen_story}
    \end{subfigure}%
    \vspace{0.5cm}
    \begin{subfigure}[t]{\textwidth}
        \includegraphics[width=\textwidth]{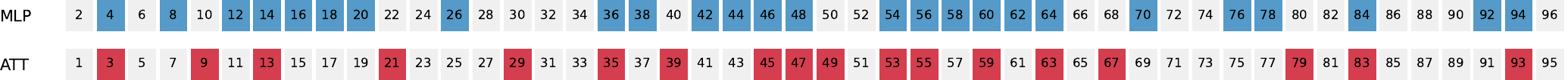}
        \caption{Text-to-SQL - Spider2}
        \label{qwen_sql}
    \end{subfigure}
    \caption{Visualization of skipped layer set configuration of Qwen-2.5-14B optimized by Self-SD \citep{zhang-etal-2024-draft} on different task domains.}
    \label{qwen_skip}
\end{figure*}

We visualize the optimal skipped layer sets we searched across five tasks on two series of models in Figure \ref{llama_skip} and Figure \ref{qwen_skip}.

\section{Datasets}
\label{datasets}

We mainly evaluate \method on LLaMA-2~\citep{touvron2023llama2openfoundation} series and Qwen-2.5~\citep{qwen2025qwen25technicalreport} series across diverse tasks, which encompass most real-life scenarios.
We select five different datasets, covering summarization, mathematical reasoning, translation, storytelling, and text-to-SQL, which are CNN/Daily Mail (CNN/DM)~\citep{nallapati-etal-2016-abstractive}, GSM8K~\citep{gsm8k}, TinyStories~\citep{tinystories}, Wmt16 DE-EN (Wmt16)~\citep{bojar-EtAl:2016:WMT1}, and Spider2~\citep{lei2025spider20evaluatinglanguage} datasets, respectively.
The maximum generation lengths on CNN/DM, GSM8K, Wmt16, Spider2, and TinyStories are set to 128, 128, 128, 128, and 512, respectively. 
We conduct 1-shot evaluation for CNN/DM and TinyStories, 3-shot evaluation for Spider2, and 5-shot evaluation for GSM8K and Wmt16.
For further analysis, we also use the XSUM~\citep{xsum-emnlp} dataset, the MATH~\citep{hendrycks2021measuringmathematicalproblemsolving} dataset, and the Alpaca~\citep{alpaca} dataset for the summarization, mathematical reasoning, and instruction following tasks, respectively.

\paragraph{CNN/DM}

The CNN/Daily Mail dataset is a large-scale benchmark for abstractive text summarization. 
It consists of long news articles paired with short summaries, derived from the CNN and Daily Mail websites. 
The dataset is used to evaluate the performance on long-form input and coherent summary generation.

\paragraph{GSM8K}

GSM8K is a high-quality benchmark dataset for arithmetic reasoning, consisting of grade school math word problems and their detailed step-by-step solutions. 
It is used to evaluate the reasoning and problem-solving capabilities in mathematical contexts.

\paragraph{TinyStories}

TinyStories is a dataset of short, synthetically generated children's stories designed to support research on language modeling and narrative understanding. 
The stories are simple in structure and vocabulary, making the dataset suitable for studying controlled text generation.

\paragraph{Wmt16}

The WMT16 De-En dataset is a standard benchmark for machine translation, consisting of parallel German-English sentence pairs collected from various sources. 
It is used to evaluate the translation quality of models.

\paragraph{Spider2}

Spider 2.0 is a complex and cross-domain text-to-SQL benchmark designed to evaluate the ability of models to generate executable SQL queries from natural language questions. 
It includes diverse databases and query types, requiring models to generalize to unseen schemas and handle intricate reasoning.

\paragraph{XSUM}

XSUM is an abstractive summarization dataset consisting of BBC news articles paired with single-sentence summaries, in contrast to CNN/DM, which provides longer, multi-sentence summaries for news articles. 
It emphasizes concise and information-rich summaries, testing the models' ability to extract key information.

\paragraph{MATH}

The MATH dataset is a benchmark for mathematical problem solving, comprising high school-level competition problems with detailed step-by-step solutions. 
It covers a wide range of topics, including algebra, counting and probability, geometry, intermediate algebra, number theory, prealgebra, and precalculus, and is designed to evaluate the advanced reasoning and symbolic manipulation abilities of language models.

\paragraph{Alpaca}

The Alpaca dataset is a collection of instruction-following demonstrations generated using the self-instruct method, based on the outputs of a strong language model. 
It covers a wide range of tasks, making it suitable for us to test the generalizability of \method.

\section{Experimental Details}

\subsection{Setups}
\label{setup}

During the pre-inference stage, we set the maximum iterations of Bayesian Optimization to 1,000 and the number of samples to 8.
For each dataset, we first randomly choose 1,000 last hidden vectors, then we use the K-means algorithm to find 10 representatives as anchors for the KNN model.

In the inference process, experiments were conducted on 8$\times$NVIDIA RTX 3090 GPU (24GB) and 4$\times$NVIDIA RTX A6000 GPU (40GB) with CUDA 12.0, and an Intel(R) Xeon(R) Gold 5117 CPU with 14 cores.
Pytorch and Huggingface transformers package are used to perform both baselines and our method.
The temperature and top\_p are set to 0.2 and 0.85, respectively.
KV cache is used during experiments.

\subsection{Evaluation Metrics}
\label{eval_metrics}

We further demonstrate the two main metrics we used in the main experiments. 
The mean accepted length $M$ denotes the average number of output tokens produced by the target LLM during each forward pass.
The token acceptance rate $\alpha$ refers to the ratio of tokens that are accepted by the target LLM to the total number of draft steps, which showcases the expectation of whether the target LLM accepts a token generated by the draft models.
Given $M$ and $\alpha$, the expected wall-time speedup can be derived as follows:
\begin{equation}
    \label{eq:eval_1}
    \mathbb{E}\text{(Speedup)} = \frac{M\alpha}{(M-1)c+\alpha}
\end{equation}
where $c$ is defined as the \textit{cost efficient} in~\citet{pmlr-v202-leviathan23a}.
It represents the ratio of the draft model's required time to the target model's during a single forward pass, which is adapted from Eq.6 in ~\citet{xia2025swift}. 
In the Self-SD method, we define $c=1-s$, where $s$ represents the proportion of skipped layers to total layers, or skip ratio, which is visualized in Appendix~\ref{pre}, as the draft model only needs to process the retained layers.

The formulation in Eq.\ref{eq:eval_1} can be made more precise by considering the difference in latency between MLP and Attention layers. 
The \textit{cost efficient} $c$ is consequently redefined as:
\begin{equation}
    \label{eq:eval_2}
    c = \frac{(1-s_{\text{mlp}})+(1-s_{\text{attn}})\times\beta}{1+\beta}
\end{equation}
where $s_{\text{mlp}}$ and $s_{\text{attn}}$ are the skipped layer ratios for MLP and Attention layers, respectively, while $\beta$ represents the latency ratio of Attention to MLP layers. 
When $\beta=1$, Eq.\ref{eq:eval_2} reduces to the original $1-s$ form.

In our experiments, using the LLaMA-2-13B model in our environment, we empirically measured the latency ratio $\beta$ to be approximately 2.3. 
For example, when the overall skipped layer ratio is $s=0.45$, and the separate skipping ratios are $s_{\text{mlp}}=0.42$ and $s_{\text{attn}}=0.48$, the corresponding values of $c$ computed using the refined formula are 0.55 and 0.538, respectively. 
These yield expected speedups of 1.434 and 1.455, a difference of only 0.021, which is not statistically significant.
Thus, we continue to use Eq.\ref{eq:eval_1} in our main experiment for simplicity.

\subsection{Details of Main Results}
\label{result_detail}

More details are provided in Table~\ref{appendix_main}.
All reported speedup results are averaged over 10 independent runs.
Results show that our \method outperforms the two Self-SD methods on both metrics, indicating our method can handle a more diverse input stream with stable inference acceleration.

\begin{table*}[ht]
  \centering
  \setlength{\tabcolsep}{4pt}
  \begin{tabular}{@{}ll
    cccccccc
    cc@{}}
    \toprule
    \multirow{2}{*}{\textbf{Models}} & \multirow{2}{*}{\textbf{Methods}}
     & \multicolumn{2}{c}{$r$=0.0} & \multicolumn{2}{c}{$r$=0.3} & \multicolumn{2}{c}{$r$=0.7} & \multicolumn{2}{c}{$r$=1.0}
     & \multicolumn{2}{c}{Overall} \\
    \cmidrule(lr){3-4} \cmidrule(lr){5-6} \cmidrule(lr){7-8} \cmidrule(lr){9-10} \cmidrule(lr){11-12}
    & & $M$ & $\alpha$ & $M$ & $\alpha$ & $M$ & $\alpha$ & $M$ & $\alpha$ & $M$ & $\alpha$ \\
    
    \midrule

    \multirow{4}{*}{LLaMA-2-13B} 
    & \texttt{Vanilla} & 1.00  & -  & 1.00 & - & 1.00 & - & 1.00 & - & 1.00 & - \\
    & \ssdf & 2.22  & 0.65  & 2.19 & 0.63 & 2.14 & 0.59 & 2.12 & 0.61 & 2.17 & 0.62 \\
    & \ssdm & 2.50  & 0.64  & 2.58 & 0.70 & 2.53 & 0.69 & 2.52 & 0.68 & 2.53 & 0.68 \\
    & \method & \textbf{3.10} & \textbf{0.86} & \textbf{3.14} & \textbf{0.88} & \textbf{3.11} & \textbf{0.89} & \textbf{3.12} & \textbf{0.88} & \textbf{3.12} & \textbf{0.88} \\
    \midrule

    \multirow{4}{*}{\makecell{LLaMA-2-13B \\ -Chat}} 
    & \texttt{Vanilla} & 1.00  & -  & 1.00 & - & 1.00 & - & 1.00 & - & 1.00 & - \\
    & \ssdf & 2.03  & 0.60 & 1.99 & 0.56 & 1.92 & 0.55 & 1.97 & 0.57 & 1.97 & 0.57 \\
    & \ssdm & 2.10  & 0.56  & 2.14 & 0.61 & 2.18 & 0.59 & 2.15 & 0.58 & 2.14 & 0.59 \\
    & \method & \textbf{2.84} & \textbf{0.84} & \textbf{2.85} & \textbf{0.86} & \textbf{2.90} & \textbf{0.85} & \textbf{2.90} & \textbf{0.86} & \textbf{2.87} & \textbf{0.85} \\

    \midrule

    \multirow{4}{*}{Qwen-2.5-14B} 
    & \texttt{Vanilla} & 1.00  & -  & 1.00 & - & 1.00 & - & 1.00 & - & 1.00 & - \\
    & \ssdf & 2.41  & 0.82  & 2.40 & 0.82 & 2.44 & 0.84 & 2.48 & 0.85 & 2.43 & 0.83 \\
    & \ssdm & 3.02  & 0.89  & 2.94 & 0.89 & 2.99 & 0.90 & 2.97 & 0.90 & 2.98 & 0.90 \\
    & \method & \textbf{4.35} & \textbf{0.99} & \textbf{4.42} & \textbf{1.00} & \textbf{4.40} & \textbf{1.00} & \textbf{4.38} & \textbf{0.99} & \textbf{4.37} & \textbf{1.00} \\

    \midrule

    \multirow{4}{*}{\makecell{Qwen-2.5-14B\\-Instruct}} 
    & \texttt{Vanilla} & 1.00  & -  & 1.00 & - & 1.00 & - & 1.00 & - & 1.00 & - \\
    & \ssdf & 2.12  & 0.80 & 2.16 & 0.80 & 2.16 & 0.80 & 2.10 & 0.79 & 2.13 & 0.80 \\
    & \ssdm & 2.32  & 0.83 & 2.25 & 0.84 & 2.35 & 0.87 & 2.34 & 0.87 & 2.32 & 0.85 \\
    & \method & \textbf{3.78} & \textbf{1.00} & \textbf{3.69} & \textbf{0.99} & \textbf{3.71} & \textbf{0.99} & \textbf{3.75} & \textbf{1.00} & \textbf{3.73} & \textbf{1.00} \\
    
    \bottomrule
  \end{tabular}  
  \caption{Comparison between \method and two Self-SD methods. $r$ indicates the mix ratio of sample streams. We report the mean accepted length and token acceptance rate, which are denoted as $M$ and $\alpha$, respectively.}
  \label{appendix_main}
\end{table*}

\end{document}